\documentclass[runningheads]{llncs}
\usepackage[T1]{fontenc}
\usepackage{graphicx,verbatim}
\usepackage[misc]{ifsym}
\usepackage{amsmath}
\usepackage{amsfonts}
\usepackage{xcolor}
\usepackage[ruled,vlined]{algorithm2e}

\newcommand{\rev}[1]{\textcolor{red}{#1}}
\newcommand{\revb}[1]{\textcolor{blue}{#1}}

\begin{document}
\title{IHF-Harmony: Multi-Modality Magnetic Resonance Images Harmonization using Invertible Hierarchy Flow Model}
\titlerunning{IHF-Harmony}

\author{Pengli Zhu$^{1}$
\and Yitao Zhu$^{1}$
\and Haowen Pang$^{1,2}$
\and Anqi Qiu$^{1,3,4}$\textsuperscript{(\Letter)}
}
\institute{
$^{1}$ Department of Health Technology and Informatics, The Hong Kong Polytechnic University, Hong Kong\\
$^{2}$ School of Integrated Circuits and Electronics, Beijing Institute of Technology, China\\
$^{3}$ Mental Health Research Center, The Hong Kong Polytechnic University, Hong Kong\\
\email{an-qi.qiu@polyu.edu.hk} \\
$^{4}$ Department of Biomedical Engineering, Johns Hopkins University, USA 
}
\authorrunning{P. Zhu et al.}

\maketitle              
\begin{abstract}
Retrospective MRI harmonization is limited by poor scalability across modalities and reliance on traveling subject datasets. To address these challenges, we introduce \textbf{IHF-Harmony}, a unified invertible hierarchy flow framework for \textit{multi-modality} harmonization \textit{using unpaired data}. By decomposing the translation process into reversible feature transformations, IHF-Harmony guarantees bijective mapping and lossless reconstruction to \textit{prevent anatomical distortion}. Specifically, an invertible hierarchy flow (IHF) performs hierarchical subtractive coupling to progressively remove artefact-related features, while an artefact-aware normalization (AAN) employs anatomy-fixed feature modulation to accurately transfer target characteristics. Combined with anatomy and artefact consistency losses, IHF-Harmony achieves high-fidelity harmonization that retains source anatomy. Experiments across multiple MRI modalities demonstrate that IHF-Harmony outperforms existing methods in both anatomical fidelity and downstream task performance, facilitating robust harmonization for large-scale multi-site imaging studies. Code is available at \url{https://github.com/Idea89560041/IHF-Harmony}.


\end{abstract}

\section{Introduction}
Magnetic resonance imaging (MRI) is a non-invasive imaging modality that provides exceptional soft-tissue contrast and is widely used for clinical diagnosis, prognosis, and treatment monitoring \cite{zhu2025t,zhu2025q}. Due to the high cost and lengthy acquisition time of MRI, large-scale neuroimaging studies increasingly rely on multi-site data collection to achieve adequate sample sizes and population diversity. 
However, aggregating data acquired from different scanners and imaging protocols inevitably introduces site-related, non-biological variability, which can confound downstream analyses and obscure subtle biological effects. Therefore, a key requirement for multi-site MRI studies is to effectively mitigate site-specific artefacts while preserving underlying anatomical structures.

Prospective harmonization attempts to reduce inter-site variability by standardizing acquisition protocols across imaging centers, but it requires extensive pilot studies, must be implemented prior to data collection, and is inapplicable to previously acquired datasets. Moreover, residual variability often persists even under harmonized protocols due to inherent differences in scanner hardware and system characteristics \cite{Shinohara2017Volumetric}. Consequently, retrospective harmonization has emerged as the preferred alternative, enabling post-acquisition correction of existing datasets and improving the reliability of downstream analyses \cite{Yu2018Statistical}.

Existing retrospective harmonization methods can be broadly categorized into statistics-based and learning-based approaches. Statistics-based methods, including intensity normalization \cite{shinohara2014,guan2022fast} and batch-effect correction \cite{Fortin2017Harmonization,Fortin2018Harmonization}, primarily operate on global intensity distributions or pre-extracted features and are limited in their ability to model spatially varying, anatomy-dependent artefacts. Learning-based methods aim to overcome this limitation by learning nonlinear mappings between imaging sites using machine learning or deep learning models \cite{ravano2022neuroimaging,cackowski2023imunity,xu2024simix,beizaee2025harmonizing}. However, supervised learning approaches typically rely on paired scans from traveling subject datasets, which are rarely available in large-scale or longitudinal studies. Unsupervised deep learning methods alleviate this requirement by learning mappings from unpaired data \cite{liu2021style,liu2024learning,wu2025unpaired,wu2026unpaired}, but most existing approaches focus on site translation between single-modality image pairs, resulting in poor scalability as the number of modalities increases and often failing to sufficiently preserve fine-grained anatomical structures.

To address the above limitations, we propose IHF-Harmony, a unified retrospective harmonization method. Unlike existing approaches, IHF-Harmony harmonizes \textit{multi-modality} MRI using a single deep learning model trained on \textit{unpaired data}. It effectively preserves anatomy by decomposing the translation into reversible feature transformations, which separate anatomical information from artefact-specific features. Our main contributions are outlined as follows:
\begin{enumerate}
	\item We propose IHF-Harmony, an invertible flow-based framework for \textit{multi-modality} MRI harmonization that jointly models anatomy and artefacts within a single network. By formulating harmonization as bijective feature transformations, the method enables \textit{lossless reconstruction} and consistent \textit{anatomy preservation} while supporting scalable multi-modality translation.
	
	\item An invertible hierarchical flow with subtractive coupling is introduced to progressively disentangle artefact-related features while preserving anatomy. This bijective formulation guarantees \textit{accurate reconstruction} and preserves \textit{fine-grained structures} that are often lost in unsupervised methods.
	
	\item We develop artefact-aware normalization, which applies anatomy-preserving transformations using learnable anatomy-guided affine parameters jointly inferred from source anatomy and target artefacts, allowing images to \textit{maintain structural identity} while \textit{adopting target artefact characteristics}.
	
	\item A comprehensive set of consistency losses is formulated to explicitly enforce \textit{structural fidelity} and \textit{realistic artefact translation}, ensuring robust performance across heterogeneous imaging protocols.
\end{enumerate}

\section{Methodology}\label{Methodology}
\subsection{Overall Architecture}
\begin{figure}[h]
	\centering
	\includegraphics[width=0.95\textwidth]{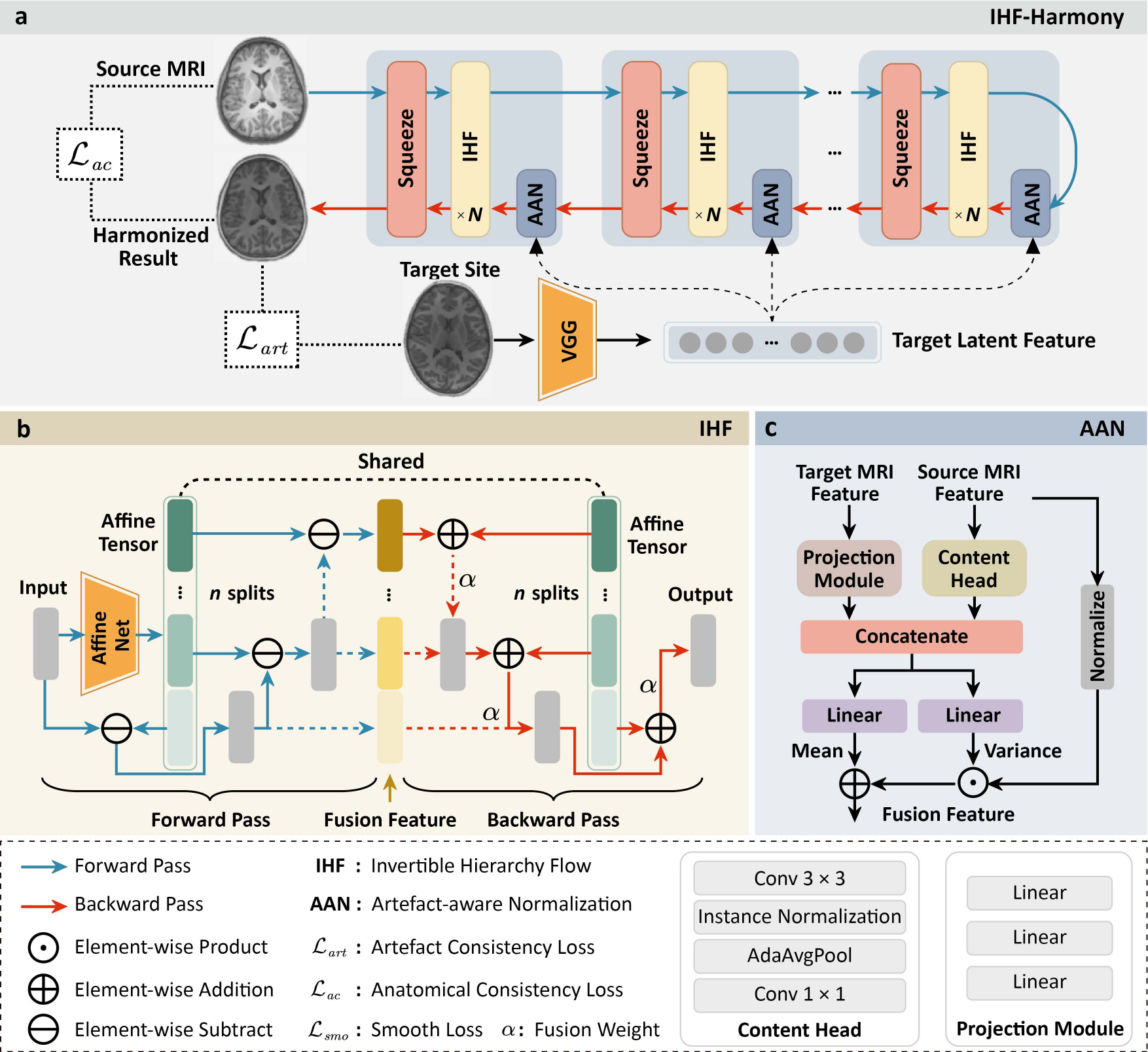}
	\vspace{-8pt}
	\caption{\textbf{Overview of IHF-Harmony.}
		\textbf{a}, IHF-Harmony works in an invertible manner. The \revb{blue arrows} indicate the forward pass for feature extraction, while the \rev{red arrows} denote the backward pass for image reconstruction. 		
		\textbf{b}, invertible hierarchy flow (IHF) performs hierarchical channel-wise subtraction to enable learnable spatial feature transformation and multi-scale fusion.		
		\textbf{c}, artefact-aware normalization (AAN) aligns source feature statistics to the target template using anatomy-guided affine parameters.}
	\label{fig:framework}
	\vspace{-20pt}
\end{figure}

Fig.~\ref{fig:framework} illustrates the IHF-Harmony framework. To explicitly disentangle anatomical structure from site-specific effects, the architecture comprises squeeze operations, invertible hierarchy flows (IHF), and artefact-aware normalization (AAN). A fixed VGG encoder $E_\text{vgg}$ is employed for feature extraction and loss computation. 
Leveraging its invertible design to prevent information loss, IHF-Harmony defines a forward pass $E$ and a backward mapping $E^{-1}$. Given a source image $\mathbf{x}$ and a target image $\mathbf{y}$, the forward pass maps $\mathbf{x}$ to latent features $\mathbf{z}$, while $E_\text{vgg}$ extracts artefact statistics $(\mu, \sigma)$ from $\mathbf{y}$. Conditioned on $(\mu, \sigma)$, the AAN module transforms $\mathbf{z}$ into $\hat{\mathbf{z}}$ in the target artefact space. The backward pass subsequently reconstructs the harmonized image $\hat{\mathbf{x}} = E^{-1}(\hat{\mathbf{z}})$, preserving the anatomy of $\mathbf{x}$ while adopting the artefact characteristics of $\mathbf{y}$. Detailed descriptions of each component and the training strategy are provided in subsequent sections.
\vspace{-15pt}

\subsection{Squeeze Operation}
To reorganize features between consecutive blocks, we adopt a squeeze operation as depicted in Fig.~\ref{fig:framework}a. 
This mechanism reduces spatial resolution by splitting the input feature into non-overlapping patches and concatenating them along the channel dimension. By compacting spatial information into channel representations, this design facilitates efficient feature interaction in subsequent modules.

\subsection{Artefact-Aware Normalization}
To perform \textit{artefact-aware feature harmonization}, we first extract artefact features from the target image as illustrated in Fig.~\ref{fig:framework}a, where IHF-Harmony decouples anatomical content into normalized features defined by a channel-wise mean $\mu$ and variance $\sigma$. The target artefact embedding $\mathbf{z}_s$ is formed by concatenating multi-scale statistics extracted from a pre-trained VGG encoder $\phi$:
\begin{equation}
	\mathbf{z}_s = \text{Concat} \left[\mu(\phi_1(\mathbf{y})), \dots, \mu(\phi_4(\mathbf{y})), \sigma(\phi_1(\mathbf{y})), \dots, \sigma(\phi_4(\mathbf{y}))\right],
\end{equation}
where $\phi_{1-4}$ denote features from $relu1\_1$ to $relu4\_1$ layers, and $\text{Concat}[\cdot]$ represents channel-wise concatenation. As depicted in Fig.~\ref{fig:framework}c, the proposed artefact-aware normalization (AAN) processes target features via a projection module and source anatomy via a content head. These representations are concatenated and fed into parallel linear networks that predict anatomy-guided affine parameters $\text{AGA}_{\sigma}$ and $\text{AGA}_{\mu}$, conditioned on both source anatomy $\mathbf{z}$ and target artefact $\mathbf{z}_s$. The resulting harmonized feature $\hat{\mathbf{z}}$ is formulated as:
\begin{equation}\label{eq:AAN}
	\hat{\mathbf{z}} = \text{AAN}(\mathbf{z}, \mathbf{z}_s) = \frac{\mathbf{z} - \mu(\mathbf{z})}{\sigma(\mathbf{z})} \cdot \text{AGA}(\mathbf{z}, \mathbf{z}_s)_\sigma + \text{AGA}(\mathbf{z}, \mathbf{z}_s)_\mu.
\end{equation}

\subsection{Invertible Hierarchy Flow (IHF)}
To achieve \textit{anatomy-preserving} harmonization with enhanced \textit{spatial fusion}, we propose an invertible hierarchy flow (IHF) via hierarchical channel-wise coupling (Fig.~\ref{fig:framework}b). The IHF operates in two stages: a subtractive forward pass for feature encoding, and an additive reverse pass for accurate reconstruction.

\noindent {\bf Forward Pass:}
Given an unharmonized image $\mathbf{x}$, we feed $\mathbf{x}$ into an affine-net to generate an affine tensor $\mathbf{a}$ with $n$-times expanded channel dimension, which is then split into $n$ channel-wise components $\{\mathbf{a}_i\}_{i=1}^{n}$.
Based on these components, a hierarchical subtractive coupling is applied iteratively. Specifically, the first intermediate feature map is computed as $\mathbf{h}_1 = \mathbf{x} - \mathbf{a}_1$, and each subsequent feature map is obtained by subtracting the corresponding affine component from the previous one, i.e., $\mathbf{h}_i = \mathbf{h}_{i-1} - \mathbf{a}_i$ for $i=2,\ldots,n$. 
Finally, the output representation $\mathbf{z}$ is formed by concatenating $\{\mathbf{h}_i\}_{i=1}^{n}$ along the channel dimension.

\noindent {\bf Reversed Pass:}
To reconstruct the harmonized image, we perform an additive coupling in $n$ steps conditioned on the affine components. 
Given the latent representation $\mathbf{z}$, we normalize it using AAN with artefact-related statistics $\mu$ and $\sigma$, yielding the transformed tensor $\mathbf{b}$, which is similarly split into $n$ channel-wise components $\{\mathbf{b}_i\}_{i=1}^{n}$.
Starting from the last split, the intermediate feature map is computed as $\mathbf{h}_n = \mathbf{a}_n + \mathbf{b}_n$. The remaining feature maps are recovered in a recursive manner by fusing the current split with the subsequent intermediate feature map, formulated as $\mathbf{h}_i = \alpha \cdot (\mathbf{a}_i + \mathbf{b}_i) + (1 - \alpha) \cdot \mathbf{h}_{i+1}$ for $i = n-1, \ldots, 1$, where $\alpha$ is a learnable fusion weight that adaptively balances the contribution of each split.
The final harmonized image $\mathbf{\hat{x}}$ is obtained as the output term $\mathbf{h}_1$.

\subsection{Loss Functions}
\vspace{-5pt}
\noindent{\textbf{Anatomical Consistency Loss.}} 
To preserve underlying anatomy, we adopt a spatially-correlative loss \cite{zhu2025cycle} based on feature self-similarity. We extract features using a pre-trained VGG network and calculate the correlation map $G_\text{ac}$ as:
\begin{equation}\label{eq:ac_loss}
	{\cal L}_\text{ac} = \left\lVert G_\text{ac}(\mathbf{x}) - G_\text{ac}(\mathbf{\hat{x}}) \right\rVert_1,
\end{equation}
where $G_\text{ac}(\mathbf{x}) = (f_{\mathbf{x}})^T(f_{\mathbf{x}})$. Here, $f_{\mathbf{x}} \in \mathbb{R}^{C \times N_p}$ represents the feature tensor of a patch with $N_p$ points and $C$ channels. This mechanism explicitly constrains the model to maintain structural consistency via point-wise correlations.

\noindent{\textbf{Artefact Consistency Loss.}} 
To address semantic misalignment that causes transfer fails in unpaired harmonization, we propose an artefact consistency loss \cite{fan2023hierarchy} that selectively matches feature distributions via channel-wise filtering.
Let $\phi_i(\cdot)$ denote the $i$-th VGG layer ($i \in \{relu1\_1, 2\_1, 3\_1\}$), we compute channel-wise discrepancy as $D = \| \mu(\phi_i(\mathbf{\hat{x}})) - \mu(\phi_i(\mathbf{y})) \|_2$, where $\mu$ denotes channel mean. 
Let $C$ index the $k \cdot N$ channels with smallest $D$ values, where $k$ is the selection ratio and $N$ is the total number of channels. The loss is defined as:
\begin{equation}
	{\cal L}_\text{art} = \sum_{i} \sum_{j \in C} \left( \| \mu(\phi_{i,j}(\mathbf{\hat{x}})) - \mu(\phi_{i,j}(\mathbf{y})) \|_2 + \| s(\phi_{i,j}(\mathbf{\hat{x}})) - s(\phi_{i,j}(\mathbf{y})) \|_2 \right),
\end{equation}
where $s$ denotes the channel standard deviation, and $k$ is empirically set to 0.9 to align artefact statistics while handling anatomical variations in unpaired data.

\noindent{\textbf{Overall Objective.}} 
The final objective is a weighted sum of the above losses:
\begin{equation}
	{\cal L}_\text{total} = \lambda_\text{ac} {\cal L}_\text{ac} + \lambda_\text{art} {\cal L}_\text{art},
\end{equation}
where $\lambda_\text{ac}$ and $\lambda_\text{art}$ are weights balancing image fidelity and harmonized intensity.

\subsection{Implementation}\label{sec:Implementation}
\vspace{-5pt}
\noindent {\bf Dataset}~
We evaluated IHF-Harmony on the Adolescent Brain Cognitive Development (ABCD) study\footnote{\url{https://abcdstudy.org/}}\cite{Jernigan2018The}, comprising multi-vendor T1, T2, and diffusion-weighted MRI. For each modality, we allocated 200 volumes per vendor for training, with approximately 2,000 volumes reserved for validation and testing. To further assess cross-site harmonization, we utilized two traveling subject datasets: SRPBS-TS\cite{tanaka2021multi} for structural MRI and HDD\cite{weninger2023deep} for diffusion MRI, allocating 5 volumes per site for training and the remainder for evaluation. All traveling images were registered to a common space prior to harmonization.

\noindent {\textbf{Optimization}}~
IHF-Harmony was implemented in PyTorch 1.13.1 and evaluated on an NVIDIA Tesla A100 GPU. To leverage volumetric context and ensure spatial smoothness, we adopted a 2.5D training strategy where each slice is stacked with two adjacent neighbors to form a three-channel input, with intensities normalized to $[-1, 1]$. We employed the ADAM optimizer\cite{kingma2014adam} with a learning rate of $1\times10^{-4}$ and a batch size of 32. The loss weights were set to $\lambda_\text{ac}=1$ and $\lambda_\text{art}=2$ across all datasets, and the training was conducted for 300 epochs.

\section{Experimental Results}\label{Study}
\subsection{Removal of Vendor Variance with Image Fidelity Preservation}
\begin{figure*}[h]
	\centering
	\includegraphics[width=0.83\textwidth]{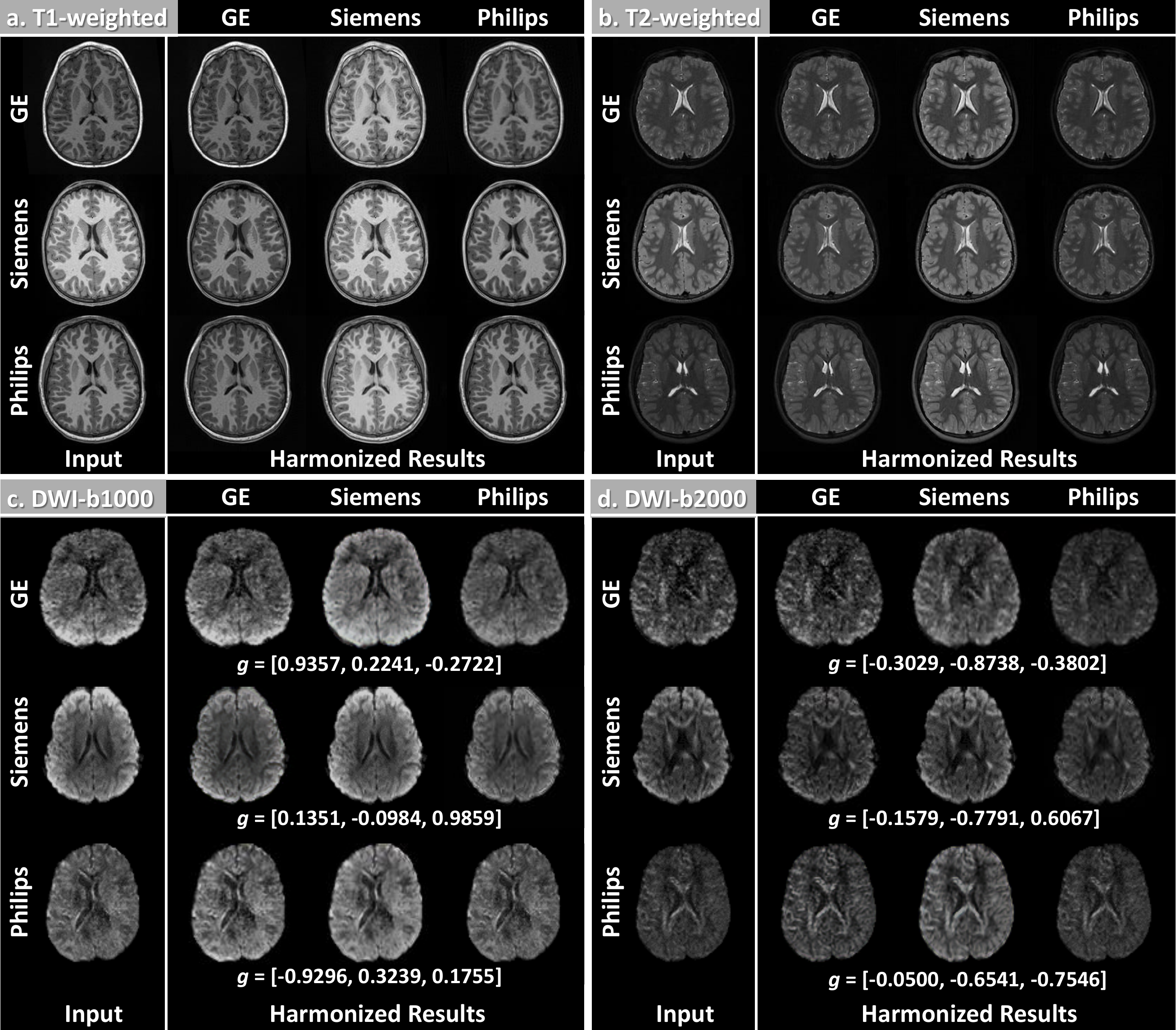}
	\vspace{-8pt}
	\caption{\textbf{Multi-modality MRI harmonization across vendors.} Representative results for \textbf{a}, T1-weighted images, \textbf{b}, T2-weighted images, and \textbf{c--d}, diffusion-weighted images. The leftmost column displays original unharmonized images, followed by results harmonized to GE, Siemens, and Philips scanners, respectively.} 
	\label{fig:gen_ref}
	\vspace{-15pt}
\end{figure*}

We evaluated the multi-vendor harmonization performance of IHF-Harmony across three MRI modalities utilizing the ABCD study \cite{Jernigan2018The}, with visualized results presented in Fig.~\ref{fig:gen_ref}. 
For each panel, the first column displays the source images, while the subsequent columns present the harmonized outputs corresponding to GE, Philips, and Siemens targets, respectively. Qualitative assessment demonstrates that IHF-Harmony effectively adapts contrast and preserves anatomical details, thereby ensuring high image fidelity. Notably, when the source and target vendors are identical, IHF-Harmony maintains both the original contrast and anatomical integrity, highlighting its robustness against over-correction.
\vspace{-15pt}
\subsection{Structural MRI Harmonization and Histogram Consistency}
\begin{figure*}[h]
	\centering
	\includegraphics[width=0.8\textwidth]{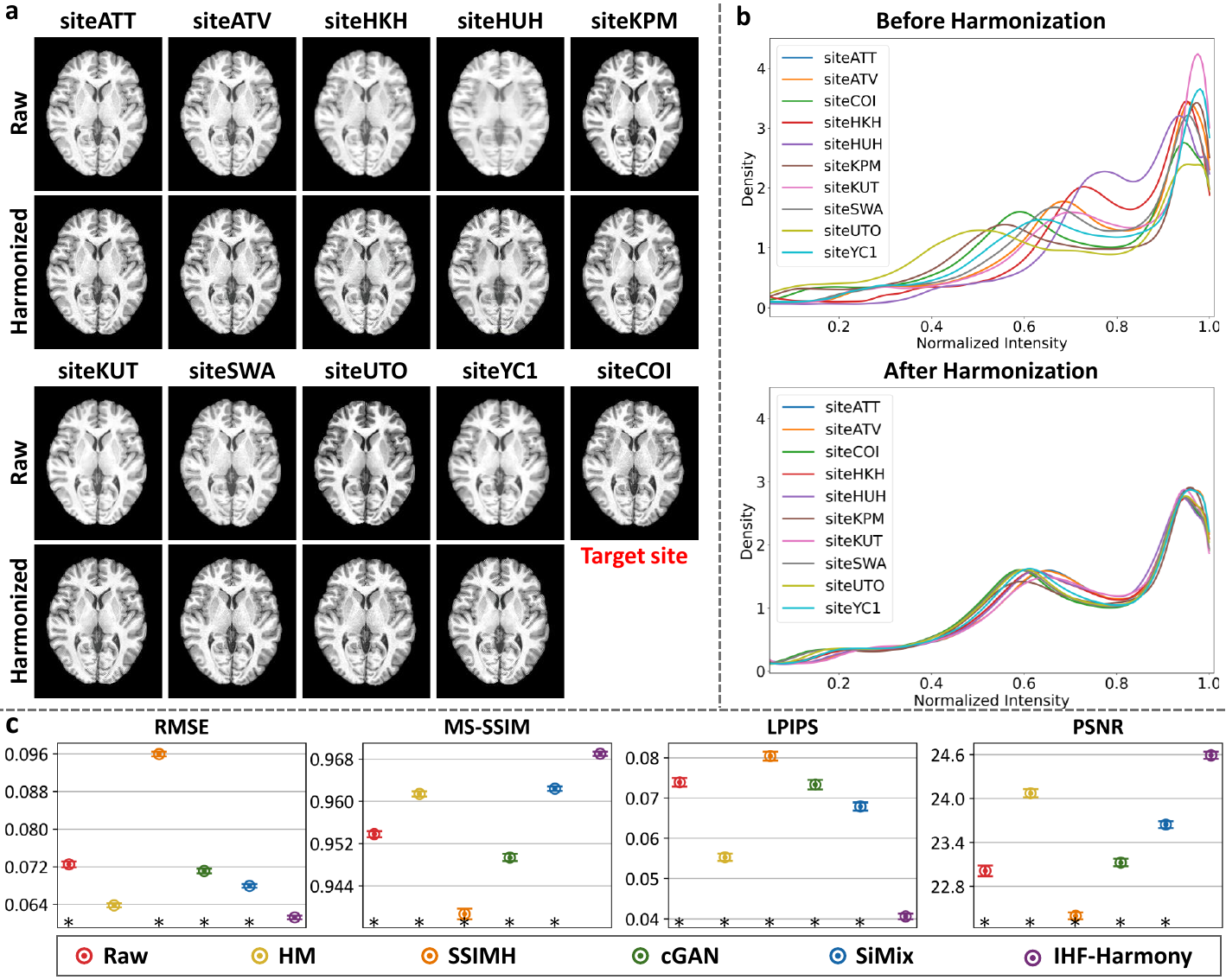}
	\vspace{-10pt}	
	\caption{\textbf{Numerical evaluation of harmonization outcomes in structural MRI.} \textbf{a--b}, visual inspection and histogram distribution comparison for multi-site harmonization. \textbf{c}, quantitative comparison of different harmonization methods. Multiple metrics were used for evaluation (*\textit{p} $<$ 0.0001, two-sided $t$-test) and mean values are marked by bullseye symbols and standard errors by error bars.}
	\label{fig:histogram}
	\vspace{-18pt}
\end{figure*}

We further evaluated IHF-Harmony on structural MRI using the SRPBS-TS dataset \cite{tanaka2021multi}, designating the COI site as the target domain. Comparisons were conducted against two non-learning baselines, including histogram matching (HM) \cite{shinohara2014} and spectrum swapping-based image-level harmonization (SSIMH) \cite{guan2022fast}, as well as two learning-based approaches, i.e., conditional GAN (cGAN) \cite{ravano2022neuroimaging} and Site Mix (SiMix) \cite{xu2024simix}. 
As shown in Fig.~\ref{fig:histogram}c, IHF-Harmony outperforms all comparative methods across multiple metrics, including RMSE, MS-SSIM, LPIPS and PSNR, indicating superior contrast harmonization and anatomical detail preservation. The visual results in Fig.~\ref{fig:histogram}a further show more consistent intensity profiles across sites while maintaining structural integrity. 
Additionally, we assessed intensity distribution consistency using kernel density estimation (KDE) \cite{silverman2018density}. As illustrated in Fig.~\ref{fig:histogram}b, the raw images exhibit pronounced, site-related intensity shifts, whereas IHF-Harmony effectively aligns the intensity distributions of all nine source sites with that of the target domain.
\vspace{-15pt}	
\vspace{-10pt}
\subsection{Diffusion MRI Harmonization and Model Fitting Consistency}
\begin{figure*}[h]
	\centering
	\includegraphics[width=0.83\textwidth]{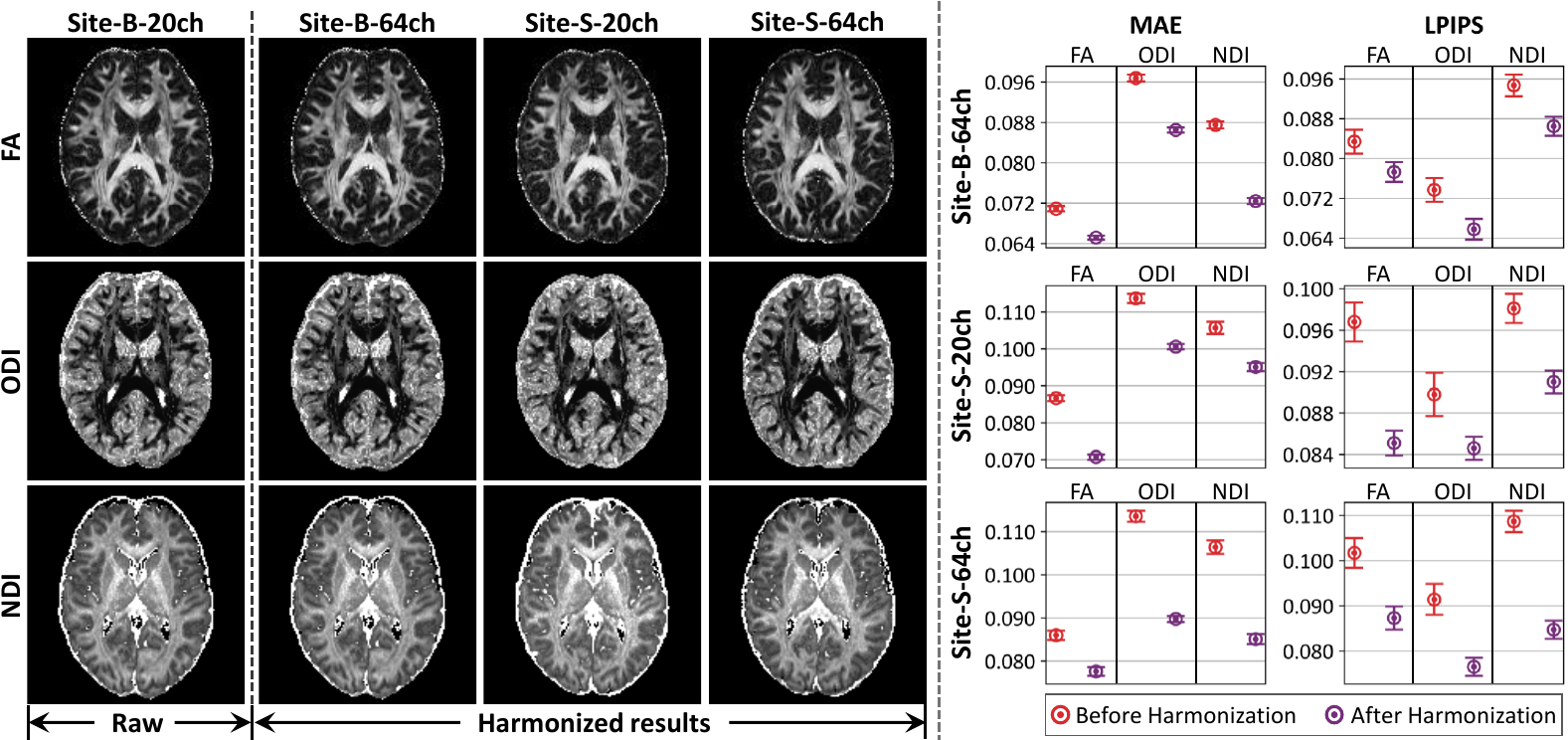}
	\vspace{-10pt}
	\caption{\textbf{Validation of diffusion MRI harmonization via parameter fitting consistency.} The left and right panels show parameter maps used for qualitative and quantitative analysis of scanner- and channel-specific harmonization results, respectively. (B/S: scanner; 20/64ch: channel)}
	\label{fig:model_fitting}
	\vspace{-30pt}
\end{figure*}

We assessed the generalizability of IHF-Harmony on diffusion MRI via the HDD dataset \cite{weninger2023deep} through downstream harmonization tasks, focusing on the consistency of three microstructural parameter maps: fractional anisotropy (FA), neurite density index (NDI), and orientation dispersion index (ODI). 
As depicted in Fig.~\ref{fig:model_fitting}, IHF-Harmony preserves complex anatomical structures while effectively mitigating scanner- and channel-induced variability. Quantitative results further demonstrate reduced site-specific bias and improved structural fidelity, confirming the effectiveness in harmonizing multi-site diffusion MRI.
\vspace{-18pt}
\subsection{Ablation Studies}
\vspace{-5pt}
To evaluate the contribution of each loss component within IHF-Harmony, we conducted ablation studies on the SRPBS-TS dataset \cite{tanaka2021multi} under identical experimental settings. As shown in Table~\ref{tb:ablation analysis}, the incremental integration of proposed loss terms leads to consistent improvements across all metrics. The decline in RMSE and LPIPS, coupled with gains in PSNR and MS-SSIM, underscores the necessity of each objective function in improving overall harmonization quality.
\vspace{-13pt}
\begin{table}[h]
	\centering
	\caption{Ablation results of IHF-Harmony with different losses}
	\vspace{-8pt}
	\renewcommand{\arraystretch}{1.5}
	\setlength{\tabcolsep}{9pt}
	\resizebox{1\textwidth}{!}{
		\begin{tabular}{c|cccc}
			\hline
			&  \textbf{RMSE $\downarrow$} & \textbf{MS-SSIM $\uparrow$} & \textbf{PSNR $\uparrow$}  & \textbf{LPIPS $\downarrow$} \\
			\hline
			\textbf{w/o ${\cal L}_\text{ac}$ \& ${\cal L}_\text{art}$}  & 0.4196 $\pm$ 0.0012 & 0.5844 $\pm$ 0.0043  & 7.54 $\pm$ 0.026 & 0.4761 $\pm$ 0.0033 \\
			\hline
			\textbf{w/o ${\cal L}_\text{art}$} & 0.1157 $\pm$ 0.0012 & 0.8844 $\pm$ 0.0012 & 18.75 $\pm$ 0.09  & 0.1416 $\pm$ 0.0014 \\
			\hline
			\textbf{w/o ${\cal L}_\text{ac}$} & 0.0935 $\pm$ 0.0007 & 0.9079 $\pm$ 0.0011 & 20.58 $\pm$ 0.06  & 0.0570 $\pm$ 0.0009 \\
			\hline
			\textbf{Full model} & \textbf{0.0612 $\pm$ 0.0004} & \textbf{0.9690 $\pm$ 0.0004} & \textbf{24.59 $\pm$ 0.05} & \textbf{0.0406 $\pm$ 0.0009} \\
			\hline
	\end{tabular}}
	\label{tb:ablation analysis}
	\vspace{-28pt}	
\end{table}
\section{Conclusion}\label{CD} 
\vspace{-7pt}	
This study presents IHF-Harmony, a unified invertible flow-based framework designed to mitigate site-specific artefacts while preserving anatomical structures in multi-modality MRI. By integrating invertible hierarchy flows with artefact-aware normalization, it enables lossless, high-fidelity harmonization using only site-labeled images, without requiring traveling subject datasets. 
Experiments demonstrate that IHF-Harmony consistently outperforms comparative methods in anatomical preservation and downstream task performance, providing a robust solution for integrating heterogeneous datasets in large-scale clinical studies.\\ 

\noindent {\bf Acknowledgements.}
This research/project is supported by Brain Science and Brain-like Intelligence Technology - National Science and Technology Major Project (2022ZD0209000). Additional support is provided by the University Grants Committee (GRF: 15201124), and the Hong Kong Global STEM Scholar scheme.\\

\noindent {\bf Disclosure of Interests.}
The authors have no competing interests to declare that are relevant to the content of this article.

\bibliographystyle{splncs04}
\bibliography{Paper}

@article{zhu2025t,
	title={Q-space Guided Multi-Modal Translation Network for Diffusion-Weighted Image Synthesis},
	author={Zhu, Pengli and Fu, Yingji and Chen, Nanguang and Qiu, Anqi},
	journal={IEEE Transactions on Medical Imaging},
	year={2025},
	publisher={IEEE}
}

@inproceedings{liu2021style,
	title={Style transfer using generative adversarial networks for multi-site MRI harmonization},
	author={Liu, Mengting and Maiti, Piyush and Thomopoulos, Sophia and Zhu, Alyssa and Chai, Yaqiong and Kim, Hosung and Jahanshad, Neda},
	booktitle={International conference on medical image computing and computer-assisted intervention},
	pages={313--322},
	year={2021},
	organization={Springer}
}

@article{xu2024simix,
	title={SiMix: A domain generalization method for cross-site brain MRI harmonization via site mixing},
	author={Xu, Chundan and Li, Jie and Wang, Yakui and Wang, Lixue and Wang, Yizhe and Zhang, Xiaofeng and Liu, Weiqi and Chen, Jingang and Vatian, Aleksandra and Gusarova, Natalia and others},
	journal={NeuroImage},
	volume={299},
	pages={120812},
	year={2024},
	publisher={Elsevier}
}

@article{wu2026unpaired,
	title = {Unpaired volumetric harmonization of brain MRI with conditional latent diffusion},
	journal = {Medical Image Analysis},
	volume = {107},
	pages = {103849},
	year = {2026},
	issn = {1361-8415},
	doi = {https://doi.org/10.1016/j.media.2025.103849},
	author = {Mengqi Wu and Minhui Yu and Shuaiming Jing and Pew-Thian Yap and Zhengwu Zhang and Mingxia Liu}
}

@inproceedings{wu2025unpaired,
	title={Unpaired Multi-site Brain MRI Harmonization with Image Style-Guided Latent Diffusion},
	author={Wu, Mengqi and Yu, Minhui and Lin, Weili and Yap, Pew-Thian and Liu, Mingxia},
	booktitle={International Conference on Medical Image Computing and Computer-Assisted Intervention},
	pages={683--693},
	year={2025},
	organization={Springer}
}

@article{beizaee2025harmonizing,
	title={Harmonizing flows: Leveraging normalizing flows for unsupervised and source-free MRI harmonization},
	author={Beizaee, Farzad and Lodygensky, Gregory A and Adamson, Chris L and Thompson, Deanne K and Cheong, Jeanie LY and Spittle, Alicia J and Anderson, Peter J and Desrosiers, Christian and Dolz, Jose},
	journal={Medical Image Analysis},
	volume={101},
	pages={103483},
	year={2025},
	publisher={Elsevier}
}

@article{liu2024learning,
	title={Learning multi-site harmonization of magnetic resonance images without traveling human phantoms},
	author={Liu, Siyuan and Yap, Pew-Thian},
	journal={Communications Engineering},
	volume={3},
	number={1},
	pages={6},
	year={2024},
	publisher={Nature Publishing Group UK London}
}

@book{silverman2018density,
	title={Density estimation for statistics and data analysis},
	author={Silverman, Bernard W},
	year={2018},
	publisher={Routledge}
}

@inproceedings{ravano2022neuroimaging,
	title={Neuroimaging harmonization using cGANs: image similarity metrics poorly predict cross-protocol volumetric consistency},
	author={Ravano, Veronica and D{\'e}monet, Jean-Fran{\c{c}}ois and Damian, Daniel and Meuli, Reto and Piredda, Gian Franco and Huelnhagen, Till and Mar{\'e}chal, B{\'e}n{\'e}dicte and Thiran, Jean-Philippe and Kober, Tobias and Richiardi, Jonas},
	booktitle={International Workshop on Machine Learning in Clinical Neuroimaging},
	pages={83--92},
	year={2022},
	organization={Springer}
}

@inproceedings{zhu2025q,
	title={Q-Space Guided Collaborative Attention Translation Network for Flexible Diffusion-Weighted Images Synthesis},
	author={Zhu, Pengli and Fu, Yingji and Chen, Nanguang and Qiu, Anqi},
	booktitle={International Conference on Medical Image Computing and Computer-Assisted Intervention},
	pages={501--511},
	year={2025},
	organization={Springer}
}

@techreport{weninger2023deep,
	title={Deep Learning-based analysis of Diffusion MRI image data from multicenter studies and from glioma patients},
	author={Weninger, Leon Dominic and M{\"u}ller, Stefan and Merhof, Dorit},
	year={2023},
	institution={Lehrstuhl f{\"u}r Bildverarbeitung}
}

@article{zhu2025cycle,
	title={Cycle-conditional diffusion model for noise correction of diffusion-weighted images using unpaired data},
	author={Zhu, Pengli and Liu, Chaoqiang and Fu, Yingji and Chen, Nanguang and Qiu, Anqi},
	journal={Medical image analysis},
	pages={103579},
	year={2025},
	publisher={Elsevier}
}

@article{fan2023hierarchy,
	title={Hierarchy Flow For High-Fidelity Image-to-Image Translation},
	author={Fan, Weichen and Chen, Jinghuan and Liu, Ziwei},
	journal={arXiv preprint arXiv:2308.06909},
	year={2023}
}

@article{shinohara2014,
	title={Statistical normalization techniques for magnetic resonance imaging},
	author={Shinohara, Russell T and Sweeney, Elizabeth M and Goldsmith, Jeff and Shiee, Navid and Mateen, Farrah J and Calabresi, Peter A and Jarso, Samson and Pham, Dzung L and Reich, Daniel S and Crainiceanu, Ciprian M and others},
	journal={NeuroImage: Clinical},
	volume={6},
	pages={9--19},
	year={2014},
	publisher={Elsevier}
}

@inproceedings{guan2022fast,
	title={Fast image-level MRI harmonization via spectrum analysis},
	author={Guan, Hao and Liu, Siyuan and Lin, Weili and Yap, Pew-Thian and Liu, Mingxia},
	booktitle={International Workshop on Machine Learning in Medical Imaging},
	pages={201--209},
	year={2022},
	organization={Springer}
}

@article{cackowski2023imunity,
	title={ImUnity: A generalizable VAE-GAN solution for multicenter MR image harmonization},
	author={Cackowski, Stenzel and Barbier, Emmanuel L and Dojat, Michel and Christen, Thomas},
	journal={Medical Image Analysis},
	volume={88},
	pages={102799},
	year={2023},
	publisher={Elsevier}
}

@article{tanaka2021multi,
	title={A multi-site, multi-disorder resting-state magnetic resonance image database},
	author={Tanaka, Saori C and Yamashita, Ayumu and Yahata, Noriaki and Itahashi, Takashi and Lisi, Giuseppe and Yamada, Takashi and Ichikawa, Naho and Takamura, Masahiro and Yoshihara, Yujiro and Kunimatsu, Akira and others},
	journal={Scientific data},
	volume={8},
	number={1},
	pages={227},
	year={2021},
	publisher={Nature Publishing Group UK London}
}

@article{kingma2014adam,
	title={Adam: A method for stochastic optimization},
	author={Kingma, Diederik P and Ba, Jimmy},
	journal={arXiv preprint arXiv:1412.6980},
	year={2014}
}

@article{Shinohara2017Volumetric,
	author = {R.T. Shinohara and J. Oh and G. Nair and P.A. Calabresi and C. Davatzikos and J. Doshi and R.G. Henry and G. Kim and K.A. Linn and N. Papinutto and D. Pelletier and D.L. Pham and D.S. Reich and W. Rooney and S. Roy and W. Stern and S. Tummala and F. Yousuf and A. Zhu and N.L. Sicotte and R. Bakshi and},
	doi = {10.3174/ajnr.a5254},
	journal = {American Journal of Neuroradiology},
	month = jun,
	number = {8},
	pages = {1501--1509},
	publisher = {American Society of Neuroradiology ({ASNR})},
	title = {Volumetric Analysis from a Harmonized Multisite Brain {MRI} Study of a Single Subject with Multiple Sclerosis},
	volume = {38},
	year = {2017}}

@article{Fortin2017Harmonization,
	author = {Jean-Philippe Fortin and Drew Parker and Birkan Tun{\c{c}} and Takanori Watanabe and Mark A. Elliott and Kosha Ruparel and David R. Roalf and Theodore D. Satterthwaite and Ruben C. Gur and Raquel E. Gur and Robert T. Schultz and Ragini Verma and Russell T. Shinohara},
	doi = {10.1016/j.neuroimage.2017.08.047},
	journal = {NeuroImage},
	month = {nov},
	pages = {149--170},
	publisher = {Elsevier {BV}},
	title = {Harmonization of multi-site diffusion tensor imaging data},
	volume = {161},
	year = {2017}}

@article{Yu2018Statistical,
	author = {Meichen Yu and Kristin A. Linn and Philip A. Cook and Mary L. Phillips and Melvin McInnis and Maurizio Fava and Madhukar H. Trivedi and Myrna M. Weissman and Russell T. Shinohara and Yvette I. Sheline},
	doi = {10.1002/hbm.24241},
	journal = {Human Brain Mapping},
	month = {jul},
	number = {11},
	pages = {4213--4227},
	publisher = {Wiley},
	title = {Statistical harmonization corrects site effects in functional connectivity measurements from multi-site {fMRI} data},
	volume = {39},
	year = {2018}}

@article{Fortin2018Harmonization,
	author = {Jean-Philippe Fortin and Nicholas Cullen and Yvette I. Sheline and Warren D. Taylor and Irem Aselcioglu and Philip A. Cook and Phil Adams and Crystal Cooper and Maurizio Fava and Patrick J. McGrath and Melvin McInnis and Mary L. Phillips and Madhukar H. Trivedi and Myrna M. Weissman and Russell T. Shinohara},
	doi = {10.1016/j.neuroimage.2017.11.024},
	journal = {NeuroImage},
	month = {feb},
	pages = {104--120},
	publisher = {Elsevier {BV}},
	title = {Harmonization of cortical thickness measurements across scanners and sites},
	volume = {167},
	year = {2018}}

@article{Jernigan2018The,
	author = {Terry L. Jernigan and Sandra A. Brown and Gayathri J. Dowling},
	doi = {10.1111/jora.12374},
	journal = {Journal of Research on Adolescence},
	month = {feb},
	number = {1},
	pages = {154--156},
	publisher = {Wiley},
	title = {The Adolescent Brain Cognitive Development Study},
	volume = {28},
	year = {2018}}
\end{document}